\documentclass{article}

\usepackage{microtype}
\usepackage{graphicx}
\usepackage{subcaption}
\usepackage{booktabs} % for professional tables
\usepackage{tabularx}

\usepackage{hyperref}

\usepackage[accepted]{icml2026}
\makeatletter
\renewcommand{\Notice@String}{\textit{Accepted at the FAGEN Workshop at the $\mathit{43}^{rd}$ International Conference on Machine Learning, Seoul, South Korea, 2026. Copyright 2026 by the author.}}
\makeatother

\usepackage{amsmath}
\usepackage{amssymb}
\usepackage{mathtools}
\usepackage{amsthm}

\usepackage[capitalize,noabbrev]{cleveref}

\theoremstyle{plain}

\theoremstyle{definition}

\theoremstyle{remark}

\usepackage[disable,textsize=tiny]{todonotes}

\newcolumntype{Y}{>{\raggedright\arraybackslash}X}

\icmltitlerunning{When Should LLMs Search?}

\begin{document}

\twocolumn[
  \icmltitle{When Should LLMs Search? \\
    Counterfactual Supervision for Search Routing}



  \begin{icmlauthorlist}
    \icmlauthor{Minho Kim}{sangmyung,dmtlabs}
  \end{icmlauthorlist}

  \icmlaffiliation{sangmyung}{Sangmyung University}
  \icmlaffiliation{dmtlabs}{DMTLABS}

  \icmlcorrespondingauthor{Minho Kim}{ilw2y123@gmail.com}

  \icmlkeywords{Machine Learning, ICML}

  \vskip 0.3in
]



\printAffiliationsAndNotice{}  % workshop notice set via \Notice@String

\begin{abstract}
  Search-augmented language models can use external evidence to compensate for limitations in parametric knowledge, but search is not uniformly beneficial: models may call search for questions they can already answer, or rely on noisy evidence when correction, clarification, or abstention would be more appropriate. We formulate this as an instance-level search-routing problem: deciding whether search is needed to improve task success relative to a no-search execution. To derive supervision, we compare no-search and forced-search outcomes for the same question and construct an oracle over $\texttt{NO\_SEARCH}$, $\texttt{SEARCH}$, and $\texttt{UNSOLVED}$ based on task-specific success. Using this oracle as both an evaluation criterion and a learning signal, we train search-routing policies with supervised fine-tuning and preference optimization, improving routing macro-F1 on oracle-eligible examples from 0.7082 to 0.8235 for Gemma E2B and from 0.7053 to 0.8365 for Qwen3.5-4B. Further analysis shows that the learned policies reduce model-specific routing failures: Gemma primarily learns no-search restraint, while Qwen further reduces missed search; residual $\texttt{UNSOLVED}$ cases reveal heterogeneous bottlenecks involving model capacity, retrieval budget, evidence use, and policy behavior.
\end{abstract}

\section{Introduction}

\begin{figure}[t]
  \centering
  \includegraphics[width=\columnwidth]{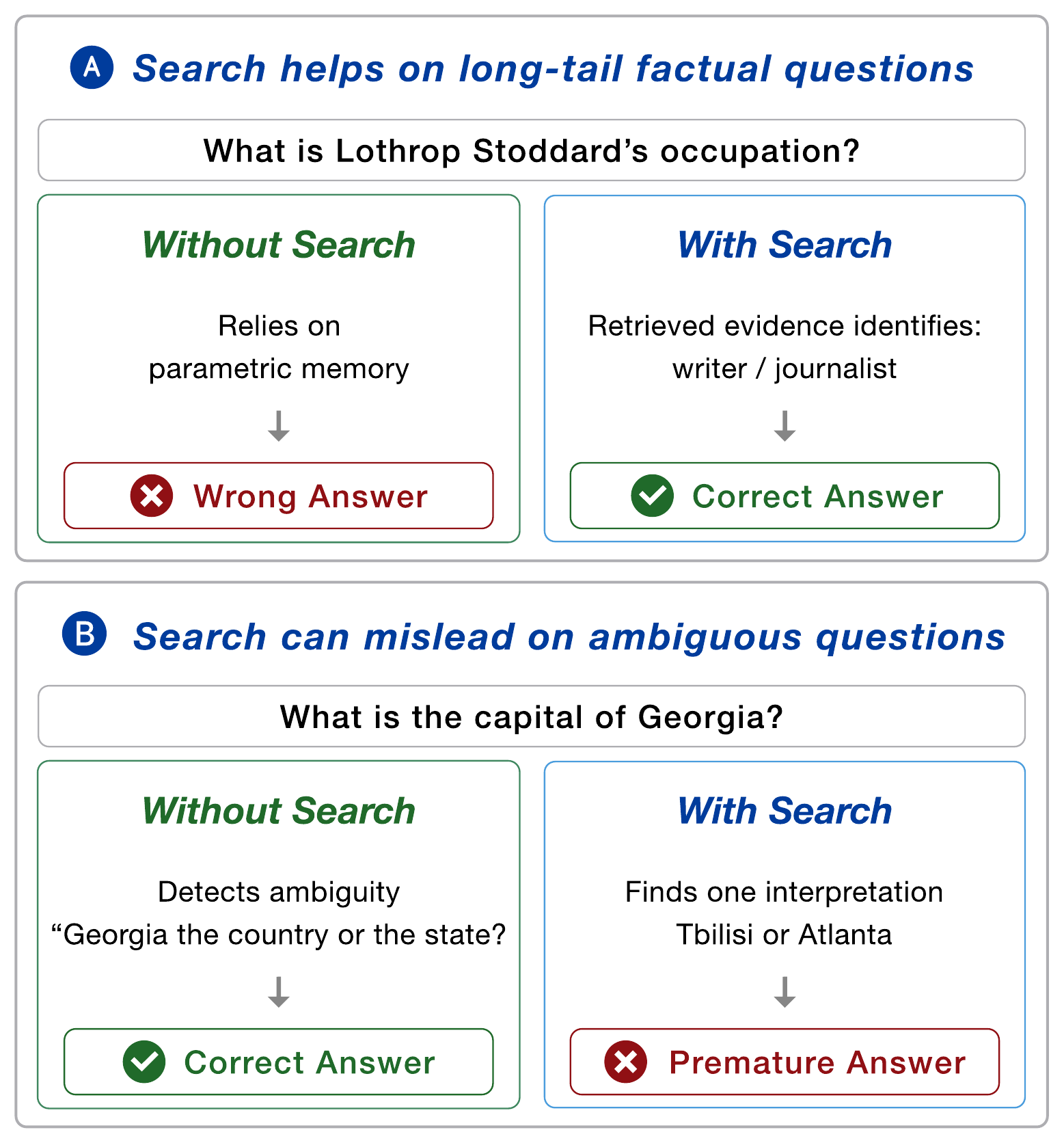}
  \caption{Motivation for instance-level search routing. Search can help, be unnecessary, or mislead, motivating instance-level routing.}
  \label{fig:main}
\end{figure}

Language models equipped with search tools can use external evidence to answer questions that are difficult to resolve from parametric knowledge alone, including queries about long-tail factual knowledge, recently updated information, or facts that are unlikely to be reliably stored in model parameters \citep{lewis_retrieval-augmented_2021,izacard_atlas_2023,mallen_when_2023}. However, the availability of a search tool does not imply that search should be used for every question. For questions the model can already answer, responding without search may be cheaper and less vulnerable to noisy retrieved evidence. For questions with false premises or missing context, the appropriate response may be to correct the premise, ask for clarification, or abstain, rather than retrieve external evidence \citep{amayuelas_knowledge_2024,xie_over-searching_2026}. In practice, the correct first action depends on the instance: a long-tail factual question may require search to recover evidence, whereas an underspecified question may be better resolved by clarification than by retrieving generic search results, as illustrated in Figure~\ref{fig:main}.

We therefore view search-augmented language models not merely as systems that generate search queries, but as systems that must first perform \emph{search routing}: deciding, for each input question, whether to respond without search or call a search tool. This first-action decision is important because routing errors arise in two opposing directions. A model that is too reluctant to search may answer directly even when its parametric knowledge is insufficient, producing plausible but incorrect responses. Conversely, a model that relies too heavily on search may invoke retrieval for questions it can already answer, or for questions that search cannot resolve, increasing cost and exposing the final answer to irrelevant or misleading evidence. The objective is not aggregate search frequency, but calling search only when it improves task success.

This framing differs from standard tool-use evaluation. Many tool-calling benchmarks focus on whether a model calls the correct tool with the correct arguments once tool use is known to be appropriate \citep{patil_berkeley_2025}. Recent work on when-not-to-call behavior broadens this view by evaluating whether models can avoid unnecessary tool calls, ask follow-up questions, or admit inability to answer \citep{ross_when2call_2025}. Our setting focuses specifically on search-augmented question answering, where the same tool may be helpful for one factual question, unnecessary for another, and actively unhelpful for a search-boundary question requiring correction, clarification, or abstention. This motivates an instance-level supervision signal that is grounded in observed response outcomes rather than in a direct judgment of whether search ``seems useful.''

We address this problem through two research questions.

\noindent\textbf{RQ1.} How accurately can a search-enabled language model identify the instances for which search improves task success relative to responding without search?

\noindent\textbf{RQ2.} Can models be trained to improve this instance-level search-routing behavior, reducing both missed search and unnecessary search?

To operationalize search need without relying on a direct human or judge annotation of search usefulness, we construct paired counterfactual traces for each question. In the no-search trace, denoted $N(q)$, the model responds without access to the search tool. In the forced-search trace, denoted $S(q)$, the model must first issue a search call and then generate a final response using the returned evidence. A task-specific evaluator maps each final response to binary success: factual correctness for answerable questions, and adequate resolution for false-premise or underspecified questions, including correction, clarification, or abstention.

We then derive a routing oracle from the paired outcomes. If the no-search trace succeeds, we assign $\texttt{NO\_SEARCH}$, since search is not necessary even if the forced-search trace also succeeds. If no-search fails but forced-search succeeds, we assign $\texttt{SEARCH}$, since search recovers a no-search failure. If both traces fail, we assign $\texttt{UNSOLVED}$. These examples do not provide a reliable binary routing target: the failure may arise from retrieval quality, query formulation, evidence use, answer synthesis, model capacity, or the question itself. We therefore exclude $\texttt{UNSOLVED}$ examples from routing training and routing-accuracy computation, while retaining them as a diagnostic subset.

This outcome-based oracle serves both evaluation and training. For evaluation, we collect a free-policy trace $P(q)$ in which the search tool is available but not forced, and compare the model's first action against the oracle label on examples labeled $\texttt{NO\_SEARCH}$ or $\texttt{SEARCH}$. This free-policy trace is a prompt-only selective-search baseline: the base model is explicitly instructed to call search only when it would materially improve the response, and otherwise to answer, correct, clarify, or abstain without using the tool. This lets us distinguish missed search, where the model answers without search despite search being necessary, from unnecessary search, where the model calls search despite no-search being sufficient. For training, we use the same oracle to construct supervised fine-tuning examples and preference pairs. Supervised fine-tuning teaches the model to imitate oracle-consistent first actions, while Preference Optimization encourages the model to prefer the appropriate first assistant completion for the same question.

Empirically, search need is model-dependent even with the same question set and search tool: parametric knowledge and search-conditioned solving ability can change whether an instance is labeled $\texttt{NO\_SEARCH}$, $\texttt{SEARCH}$, or $\texttt{UNSOLVED}$. Prompted base models nevertheless exhibit both under-search and over-search, showing that selective-search instructions alone do not recover this boundary. Training on model-specific counterfactual outcomes improves routing: on held-out oracle-eligible examples, SFT and Preference Optimization raise macro-F1 from 0.7082 to 0.8235 for Gemma E2B and from 0.7053 to 0.8365 for Qwen3.5-4B; error-direction and $\texttt{UNSOLVED}$ analyses separate corrected routing errors from residual bottlenecks beyond first-action routing.

\begin{figure*}[t]
  \centering
  \includegraphics[width=\textwidth]{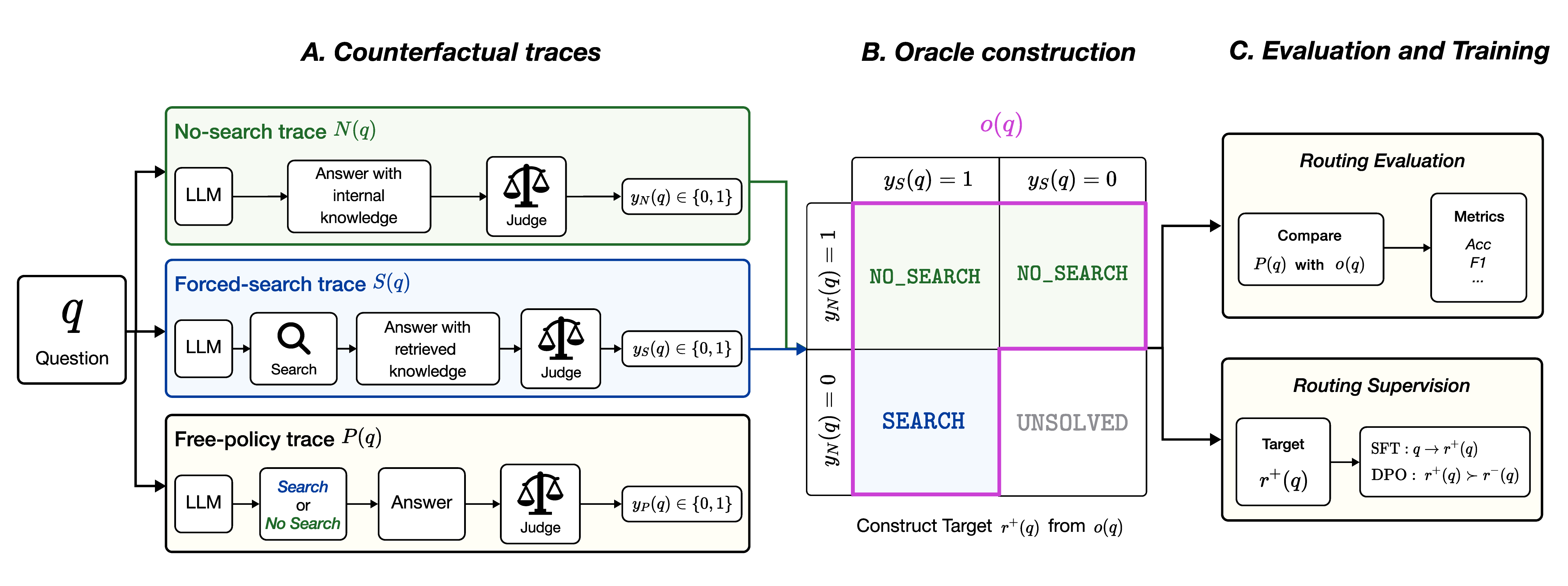}
  \caption{Counterfactual search-routing pipeline. (A) We collect controlled no-search and forced-search traces, together with a free-policy trace collected under a selective-search prompt. (B) Paired no-search and forced-search outcomes induce a routing oracle over $\texttt{NO\_SEARCH}$, $\texttt{SEARCH}$, and $\texttt{UNSOLVED}$. (C) The oracle is used both for routing supervision and for free-policy routing evaluation.}
  \label{fig:pipeline}
\end{figure*}

These findings rest on three contributions.

\begin{itemize}
  \item \textbf{Instance-level routing.} We frame search use as an instance-level $\texttt{NO\_SEARCH}/\texttt{SEARCH}$ decision, where $\texttt{NO\_SEARCH}$ includes direct answering, premise correction, clarification, and abstention.
  \item \textbf{Counterfactual oracle.} We derive an outcome-based oracle from paired no-search and forced-search traces and use it for both routing evaluation and training.
  \item \textbf{$\texttt{UNSOLVED}$ diagnostics.} We retain dual-failure cases as a diagnostic subset, showing residual retrieval, evidence-use, synthesis, and model-capacity bottlenecks beyond the initial search decision.
\end{itemize}

\section{Method}

We formalize search routing as a first-action decision. Given a question $q$, the policy selects $a(q)\in\{\texttt{NO\_SEARCH},\texttt{SEARCH}\}$, where $\texttt{NO\_SEARCH}$ denotes a no-tool first response and $\texttt{SEARCH}$ denotes a first-turn search call. Figure~\ref{fig:pipeline} gives an overview; below we define the traces, oracle, evaluation target, and training objectives.

For each question $q$, we collect two controlled counterfactual traces: $N(q)$, generated with search disabled, and $S(q)$, generated under a forced first-turn search call. We evaluate only the final response in each trace:
\[
  y_N(q)=\mathrm{Eval}(q,N(q)), \qquad
  y_S(q)=\mathrm{Eval}(q,S(q)),
\]
where $y_N,y_S\in\{0,1\}$. $\mathrm{Eval}$ applies task-specific success criteria; details and prompt templates are in Appendices~A and~C.

We derive a routing oracle from the paired outcomes:
\[
  o(q)=
  \begin{cases}
    \texttt{NO\_SEARCH}, & y_N(q)=1,\\
    \texttt{SEARCH}, & y_N(q)=0 \land y_S(q)=1,\\
    \texttt{UNSOLVED}, & y_N(q)=0 \land y_S(q)=0.
  \end{cases}
\]
Rows with $o(q)=\texttt{UNSOLVED}$ are excluded from routing supervision and routing metrics because paired failures do not identify a reliable first-action target. We retain them for diagnostic analysis in Section~5 and Appendix~B. Table~\ref{tab:oracle} summarizes the induced buckets and labels.

\begin{table}[t]
  \centering
  \small
  \caption{Oracle labels induced by counterfactual no-search and forced-search outcomes.}
  \label{tab:oracle}
  \begin{tabular}{@{}c c c l@{}}
    \toprule
    $y_N(q)$ & $y_S(q)$ & Bucket & Oracle \\
    \midrule
    1 & 1 & $N1\_S1$ & $\texttt{NO\_SEARCH}$ \\
    1 & 0 & $N1\_S0$ & $\texttt{NO\_SEARCH}$ \\
    0 & 1 & $N0\_S1$ & $\texttt{SEARCH}$ \\
    0 & 0 & $N0\_S0$ & $\texttt{UNSOLVED}$ \\
    \bottomrule
  \end{tabular}
\end{table}

Because the no-search and forced-search outcomes can differ across models,
the induced oracle is model-specific rather than a fixed question-level
annotation. Supervision therefore targets the current model's boundary between
cases it can solve without search and cases recovered by search. If no-search
already succeeds, the instance is labeled $\texttt{NO\_SEARCH}$ even when
forced search also succeeds, so $\texttt{SEARCH}$ labels mark only observed
no-search-to-search recoveries.

For free-policy evaluation, we collect $P(q)$ under a selective-search prompt: search is available but should be called only when it materially improves the response; otherwise the model should answer, correct, clarify, or abstain without the tool. We parse the first action as $p(q)=\texttt{SEARCH}$ if it calls search and $p(q)=\texttt{NO\_SEARCH}$ otherwise. When collected from the base model before routing training, this trace is denoted $P_{\mathrm{pre}}(q)$, and the corresponding prompted base routing policy is $P_{\mathrm{pre}}$. Routing evaluation measures agreement between $p(q)$ and $o(q)$ on examples with $o(q)\in\{\texttt{NO\_SEARCH},\texttt{SEARCH}\}$. We call $o(q)=\texttt{SEARCH},p(q)=\texttt{NO\_SEARCH}$ a missed search, and $o(q)=\texttt{NO\_SEARCH},p(q)=\texttt{SEARCH}$ an unnecessary search. Generation details and prompts are in Appendices~A and~C.

The free-policy prompt and search interface are held fixed for $P_{\mathrm{pre}}$ collection, training inputs, and post-training rollouts, so comparisons among $P_{\mathrm{pre}}$, SFT, and Preference Optimization reflect learned routing changes rather than instruction changes. SFT uses only oracle-eligible examples: $\texttt{NO\_SEARCH}$ targets are successful $N(q)$ responses, and $\texttt{SEARCH}$ targets are first-turn search calls from successful $S(q)$ traces, focusing supervision on the routing action rather than the full post-search trajectory.

Thus, the $\texttt{SEARCH}$-labeled SFT examples do not require the model to imitate an entire searched answer trajectory. They supervise the decision to enter the search path, while leaving query refinement, evidence use, and final synthesis outside the main routing objective. This choice is consistent with our evaluation, which scores the first action rather than the quality of all subsequent search-conditioned steps.

For Preference Optimization, each eligible $q$ has a chosen response $r^+(q)$ beginning with the oracle-consistent first action and a rejected response $r^-(q)$ beginning with the opposite action, preferably drawn from an actual $P_{\mathrm{pre}}$ mistake and otherwise from the controlled counterfactual trace.

We optimize a DPO objective with chosen-response NLL regularization \citep{rafailov_direct_2023,pang_iterative_2024}:
\[
  \mathcal{L}
  =
  \mathcal{L}_{\mathrm{DPO}}
  +
  \alpha \, \mathcal{L}_{\mathrm{NLL}}(r^+).
\]
Preference Optimization is initialized from the selected SFT checkpoint and uses that checkpoint as the reference model.

\section{Experimental Setup}

\noindent\textbf{Data and evaluation.} We instantiate the counterfactual routing protocol on two complementary question regimes. PopQA provides answerable entity-centric factual questions, for which search may recover long-tail facts; we use all 13,763 questions \citep{mallen_when_2023}. KUQ provides search-boundary questions for which search may or may not be useful: some cases may benefit from external evidence, while others are better resolved through premise correction, clarification, or abstention \citep{amayuelas_knowledge_2024}. We use the False Assumption category, where the model should reject or correct a false premise, and the Ambiguous category, where the model should recognize missing context and ask for clarification or abstain. The KUQ subset contains 774 False Assumption questions and 850 Ambiguous questions.

Importantly, these dataset groups are not treated as fixed routing labels. PopQA questions are not assumed to require search: if the no-search trace answers correctly, the instance is labeled \texttt{NO\_SEARCH}. Likewise, KUQ false-assumption or ambiguous questions are not assumed to forbid search: if no-search fails but forced search enables the model to resolve the premise or ambiguity, the instance is labeled \texttt{SEARCH}. The mixed benchmark therefore tests instance-level routing rather than dataset-level classification.

PopQA is evaluated with a PopQA-adapted SimpleQA-style factuality judge that performs alias-aware correctness checking \citep{wei_measuring_2024}, while KUQ is evaluated by category-specific resolution adequacy. Evaluator settings and binary success mappings are provided in Appendix~A; full judge prompts are provided in Appendix~C.

\noindent\textbf{Models and search environment.} We evaluate Qwen3.5-4B \citep{qwen3.5} and Gemma E2B \citep{google2026gemma4}, comparing $P_{\mathrm{pre}}$, SFT, and Preference Optimization on the test split. $P_{\mathrm{pre}}$ is the base-model rollout under the same selective-search prompt and $\texttt{search(query)}$ interface used for post-training evaluation. The interface is backed by Brave LLM Context and returns source-backed snippets. We fix the backend, interface, and routing instruction to isolate first-action search routing; budgets and forced-search drop handling are in Appendix~A.

\noindent\textbf{Trace collection.} For each question, we collect the three traces defined in Section~2. The no-search trace disables tools; the forced-search trace uses prompt-level forcing and acceptance filtering for a valid first-turn search call; and the free-policy trace uses the fixed selective-search prompt without first-action forcing. Retry/drop handling and prompt templates are in Appendices~A and~C.

\noindent\textbf{Training and metrics.}
We split data into train, dev, and test before training. SFT and Preference Optimization use only oracle-eligible train rows, formatted with the same free-policy prompt used for $P_{\mathrm{pre}}$ and post-training evaluation. \(\texttt{UNSOLVED}\) rows are excluded from training and routing metrics but retained for diagnostics. Dev selects checkpoints; test is used only for final reporting. Because oracle labels are model-specific, comparisons are primarily within model across policies rather than direct cross-model rankings.

On oracle-eligible examples, we report routing accuracy and macro-F1, the unweighted average of \(\texttt{SEARCH}\) and \(\texttt{NO\_SEARCH}\) F1. Directional error rates use \(\texttt{SEARCH}\) as the positive class, so false positives are unnecessary search and false negatives are missed search. Over-search, or unnecessary-search rate, is
\[
\mathrm{OverSearch}
=
\frac{FP}{TN+FP},
\]
the false-positive rate conditioned on oracle \(\texttt{NO\_SEARCH}\) examples. Under-search, or missed-search rate, is
\[
\mathrm{UnderSearch}
=
\frac{FN}{FN+TP},
\]
the false-negative rate conditioned on oracle \(\texttt{SEARCH}\) examples.

For category-level reporting, accuracy is computed within each dataset/category on oracle-eligible examples only; FA Acc. and Amb. Acc. are KUQ routing accuracies, not final-response accuracies. Split sizes, conditional-error denominators, and hyperparameters are in Appendix~A.

\begin{table*}[t]
  \centering
  \small
  \caption{Main search-routing results on oracle-eligible test examples. Over-search and under-search are conditional error rates.}
  \label{tab:main-routing-results}
  \resizebox{\textwidth}{!}{%
  \begin{tabular}{@{}l l c c c c c c@{}}
    \toprule
    Model & Policy & Macro-F1 & Routing Acc. & Cond. Over-search & Cond. Under-search & FA Acc. & Amb. Acc. \\
    \midrule
    Gemma E2B & $P_{\mathrm{pre}}$ & 0.7082 & 0.7574 & 0.5425 & \textbf{0.0598} & 0.6897 & 0.7170 \\
    Gemma E2B & SFT & 0.8207 & 0.8292 & \textbf{0.1928} & 0.1574 & 0.7759 & 0.8868 \\
    Gemma E2B & Preference Opt. & \textbf{0.8235} & \textbf{0.8329} & 0.2059 & 0.1434 & \textbf{0.8103} & \textbf{0.9434} \\
    Qwen3.5-4B & $P_{\mathrm{pre}}$ & 0.7053 & 0.7148 & 0.3649 & 0.2272 & 0.8732 & 0.6531 \\
    Qwen3.5-4B & SFT & 0.8207 & 0.8263 & \textbf{0.2284} & 0.1339 & \textbf{0.9296} & 0.8776 \\
    Qwen3.5-4B & Preference Opt. & \textbf{0.8365} & \textbf{0.8427} & 0.2312 & \textbf{0.1034} & \textbf{0.9296} & \textbf{0.9184} \\
    \bottomrule
  \end{tabular}%
  }
\end{table*}

\section{Results}

This section answers the two research questions. We first compare the counterfactual outcomes of no-search and forced-search traces, showing that the effect of search depends strongly on the question type and instance. We then analyze the under-search and over-search errors exhibited by the base free-policy execution $P_{\mathrm{pre}}$, and evaluate whether SFT and Preference Optimization improve first-action search routing.

\subsection{Counterfactual Outcomes Show Instance-Dependent Search Utility}

We first compare task success between controlled no-search and forced-search traces. Table~\ref{tab:counterfactual-outcomes} shows that search has opposite effects across question groups. In PopQA, forced search achieves substantially higher success rates than no search for both models. Qwen3.5-4B improves from 0.1831 under no search to 0.5096 under forced search, while Gemma E2B improves from 0.1504 to 0.4813. PopQA therefore supplies many cases in which search is beneficial: factual failures that are not resolved without search can often be recovered when the model sees external evidence.

\begin{table}[h]
  \centering
  \small
  \caption{Counterfactual search outcomes. Task success under no-search and forced-search conditions.}
  \label{tab:counterfactual-outcomes}
  \resizebox{\columnwidth}{!}{%
  \begin{tabular}{@{}l l c c@{}}
    \toprule
    Model & Question group & No-search & Forced-search \\
    \midrule
    Qwen3.5-4B & PopQA & 0.1831 & 0.5096 \\
    Gemma E2B & PopQA & 0.1504 & 0.4813 \\
    Qwen3.5-4B & KUQ FA + Amb. & 0.6441 & 0.5376 \\
    Gemma E2B & KUQ FA + Amb. & 0.6022 & 0.4483 \\
    \bottomrule
  \end{tabular}%
  }
\end{table}

By contrast, for KUQ False Assumption and Ambiguous questions, forced search yields lower success rates than no search. Qwen3.5-4B decreases from 0.6441 to 0.5376, and Gemma E2B decreases from 0.6022 to 0.4483. For these questions, no-search behaviors such as correcting a false premise, requesting missing information, or abstaining may satisfy the task-specific success criterion more reliably than searching for additional evidence.

The oracle-bucket distributions in Tables~\ref{tab:popqa-oracle-buckets} and~\ref{tab:kuq-oracle-buckets} (Appendix~A) further show that search utility varies by instance across both PopQA and KUQ search-boundary questions.

\subsection{SFT and Preference Optimization Improve Search Routing}

Table~\ref{tab:main-routing-results} compares the search-routing performance of $P_{\mathrm{pre}}$, SFT, and Preference Optimization on the test split. Evaluation is conducted on oracle-eligible test examples, i.e., examples whose oracle label is either $\texttt{NO\_SEARCH}$ or $\texttt{SEARCH}$. We report routing macro-F1, routing accuracy, conditional over-search, conditional under-search, and subset routing accuracies for KUQ False Assumption and Ambiguous questions.

Because $P_{\mathrm{pre}}$, SFT, and Preference Optimization are all evaluated under the same selective-search prompt and search interface, these comparisons do not conflate training gains with prompt changes. Moreover, $P_{\mathrm{pre}}$ is a prompt-only selective-search baseline rather than an unconstrained tool-available rollout: the base model is already instructed to use search only when it would materially help and to avoid unnecessary tool use. The remaining $P_{\mathrm{pre}}$ errors therefore reflect failures to apply this selective-use instruction at the instance level.

SFT substantially improves routing performance for both models beyond this prompt-only baseline. Gemma E2B improves from a $P_{\mathrm{pre}}$ macro-F1 of 0.7082 and routing accuracy of 0.7574 to 0.8207 and 0.8292 after SFT. Qwen3.5-4B similarly improves from a $P_{\mathrm{pre}}$ macro-F1 of 0.7053 and routing accuracy of 0.7148 to 0.8207 and 0.8263 after SFT. The counterfactual oracle therefore provides a usable learning signal for sharpening the first-action routing boundary beyond prompt-only selective search.

Preference Optimization further shifts the routing boundary after SFT. For Qwen3.5-4B, macro-F1 increases from 0.8207 to 0.8365 relative to SFT, and routing accuracy increases from 0.8263 to 0.8427. The conditional under-search rate decreases from 0.1339 to 0.1034, further reducing cases where the model omits search for questions that require it. For Gemma E2B, the aggregate gain is smaller, but macro-F1 increases from 0.8207 to 0.8235 and routing accuracy from 0.8292 to 0.8329. Preference Optimization also improves the KUQ search-boundary subsets, especially Ambiguous, where accuracy increases from 0.8868 to 0.9434.

\subsection{Error Directions Differ Across Model Families}

The error directions in Table~\ref{tab:main-routing-results} show that the two models have different initial search-routing tendencies. Gemma E2B exhibits pronounced over-search on $\texttt{NO\_SEARCH}$ oracle rows under $P_{\mathrm{pre}}$. SFT greatly reduces this over-search, although it introduces a higher under-search rate. Preference Optimization does not completely remove this trade-off, but it slightly reduces under-search and yields additional gains on the KUQ subsets, especially Ambiguous.

By contrast, Qwen3.5-4B exhibits both over-search and under-search before training, and SFT mitigates both error types. Preference Optimization further reduces under-search while keeping over-search nearly unchanged, suggesting that it makes the policy more willing to search on $\texttt{SEARCH}$-oracle cases without substantially increasing unnecessary search on $\texttt{NO\_SEARCH}$-oracle cases.

A single global search-rate target would fail for both models. For a search-heavy model such as Gemma E2B, learning no-search restraint is important; for a model such as Qwen3.5-4B, which also misses questions requiring search, repairing missed search is important. Search-routing training should therefore not simply increase or decrease the search-call rate globally. Instead, it should jointly correct under-search and over-search according to each model's initial routing tendency.

\section{Diagnostic Analysis: Heterogeneous Failures in $\texttt{UNSOLVED}$ Cases}

The oracle intentionally excludes $\texttt{UNSOLVED}$ examples from routing training and routing evaluation. These are questions for which both the no-search and forced-search traces fail, $y_N(q)=0$ and $y_S(q)=0$. Such failures do not identify a reliable first-action target. A failed no-search trace may reflect insufficient parametric knowledge or generation capacity, while a failed forced-search trace may reflect query formulation, retrieval coverage, evidence use, answer synthesis, or model capacity \citep{krishna_fact_2025}. We therefore treat $\texttt{UNSOLVED}$ as a diagnostic category rather than as either a $\texttt{NO\_SEARCH}$ or $\texttt{SEARCH}$ routing label.

To focus on persistent controlled-trace failures, we additionally evaluate the base selective-search free-policy response $P_{\mathrm{pre}}(q)$. Cases in which $y_N(q)=0$, $y_S(q)=0$, but $P_{\mathrm{pre}}$ succeeds are assigned to an escape subset. We focus on the hard subset: examples for which the no-search trace, forced-search trace, and base free-policy trace all fail.

On this hard subset, we test four diagnostic probes: a larger model without search $M_N$, a larger model with forced search $M_S$, an expanded retrieval-budget condition using the same base model $R_{\mathrm{TR}}$, and the selected post-training policy rollout $P_{\mathrm{POST}}$. For Qwen3.5-4B, the larger diagnostic model used in $M_N$ and $M_S$ is Qwen3.5-9B \citep{qwen3.5}; for Gemma E2B, it is Gemma E4B \citep{google2026gemma4}. These probes are diagnostic signatures rather than independent causal factors. For example, $M_S$ jointly changes model capacity, query formulation, evidence use, and answer synthesis. We therefore examine which probe uniquely rescues each hard case.

\begin{table}[t]
  \centering
  \small
  \caption{Exact-only rescue signatures on $\texttt{N0\_S0}$ hard subsets. Each column reports examples solved only under that diagnostic probe while the other three probes fail. Examples rescued by multiple probes are excluded. Entries show rescue rate within each model's hard subset.}
  \label{tab:diagnostic-rescue}
  \resizebox{\columnwidth}{!}{%
  \begin{tabular}{@{}l c c c c@{}}
    \toprule
    Model & $M_N$ only & $M_S$ only & $R_{\mathrm{TR}}$ only & $P_{\mathrm{POST}}$ only \\
    \midrule
    Qwen3.5-4B & 0.0202 & 0.0527 & 0.0299 & 0.0255 \\
    Gemma E2B & 0.0180 & 0.0445 & 0.0156 & 0.0211 \\
    \bottomrule
  \end{tabular}%
  }
\end{table}

Table~\ref{tab:diagnostic-rescue} reports the exact-only rescue signatures. These exclusive rescue patterns show that the hard $\texttt{N0\_S0}$ subset is not a single failure type. Some cases are recovered only by larger-model no-search inference, suggesting failures associated with parametric knowledge or no-search generation capacity. A larger share is recovered only by larger-model forced search, suggesting failures that require search-conditioned larger-model inference rather than larger no-search inference alone. Expanded retrieval also uniquely recovers a nontrivial share of cases, indicating that retrieval budget or search-context coverage can be the limiting factor even when the base forced-search trace fails. Finally, post-training policy rollout uniquely recovers some hard cases, showing that routing-related behavior can coexist with other sources of failure even within the $\texttt{N0\_S0}$ region.

These exclusive rescue signatures should not be read as a causal decomposition of mutually exclusive error sources. Instead, they show why collapsing $\texttt{N0\_S0}$ into either $\texttt{NO\_SEARCH}$ or $\texttt{SEARCH}$ would be misleading: the bucket contains a mixture of model-capacity limitations, search-conditioned synthesis failures, retrieval-budget failures, and policy-related failures. $\texttt{UNSOLVED}$ is therefore better treated as a separate diagnostic category for failures beyond reliable first-action routing supervision.

\section{Related Work}

\noindent\textbf{Adaptive retrieval and selective search.} Retrieval augmentation is a standard way to compensate for limitations in LLMs' parametric knowledge \citep{lewis_retrieval-augmented_2021,izacard_atlas_2023}. Prior work on long-tail factual QA shows that language models struggle to recall less frequent entity knowledge, and that retrieval can substantially improve performance on such questions \citep{mallen_when_2023}. However, retrieval is not uniformly beneficial: for questions the model can already answer, retrieved context may instead distract the model or degrade the response \citep{mallen_when_2023,wang_self-knowledge_2023}. Self-knowledge-guided retrieval is similarly motivated by the observation that retrieved knowledge is not always helpful, and proposes selective retrieval based on whether a model appears to know the answer \citep{wang_self-knowledge_2023}. Other adaptive retrieval systems trigger retrieval during generation or select among retrieval strategies using confidence, reflection, or question-complexity signals \citep{jiang_active_2023,asai_self-rag_2023,jeong_adaptive-rag_2024}. Our work shares the view that search should be used selectively, but differs in how the routing signal is constructed. Rather than predicting search necessity with a separate classifier, heuristic, or self-knowledge prompt, we compare no-search and forced-search task outcomes for the same question and derive instance-level routing supervision from observed success and failure.

\noindent\textbf{Tool calling and when-not-to-call behavior.} Tool-use and tool-calling work primarily evaluates whether models can generate the correct tool name, arguments, and executable function calls \citep{schick_toolformer_2023,li_api-bank_2023,qin_toolllm_2024,zhang_toolbehonest_2024,lu_toolsandbox_2025,patil_berkeley_2025}. BFCL is a representative benchmark that evaluates function-calling accuracy across single-turn, multi-turn, and agentic settings \citep{patil_berkeley_2025}. When2Call moves closer to the question of when tools should not be called by asking models to choose among tool calling, direct answering, follow-up questioning, and being unable to answer \citep{ross_when2call_2025}. However, When2Call is formulated as a multiple-choice benchmark, and its questions are often structured so that direct answering is an inappropriate or hallucination-prone option. By contrast, $\texttt{NO\_SEARCH}$ in our setting is not defined as a failure behavior. In PopQA, it may correspond to a correct factual answer produced without search; in KUQ-style boundary cases, it may correspond to premise correction, clarification, or abstention. Thus, our work learns a model-specific search-routing boundary from free-form execution traces, based on whether an actual $\texttt{search(query)}$ execution recovers task success.

\noindent\textbf{Over-searching and uncertainty-aware no-search behavior.} Recent work on over-searching shows that search can improve answer accuracy for answerable queries while reducing abstention accuracy and increasing cost for unanswerable or underspecified queries \citep{xie_over-searching_2026}. Such work highlights the risks of unnecessary search, especially when retrieved evidence is noisy or when the question should instead be corrected, clarified, or rejected. This also relates to work on aligning assistants to refuse or say ``I don't know'' on model-specific unknown questions \citep{cheng_can_2024}. Our work shares this concern, but does not focus only on over-search. We jointly address over-search, where the model calls search for questions where it is unnecessary, and under-search, where the model omits search for questions where search would recover a no-search failure. In addition, we convert observed no-search and forced-search outcomes into supervision for SFT and Preference Optimization, directly training policies to correct bidirectional search-routing errors.

\noindent\textbf{Known-unknown questions and response-level evaluation.} Known-unknown question datasets such as KUQ categorize sources of uncertainty, including false assumptions and ambiguous or underspecified questions \citep{amayuelas_knowledge_2024}. These categories are useful for our setting because they expose cases in which the value of search is itself uncertain. A false-premise or ambiguous question is not automatically a $\texttt{NO\_SEARCH}$ case: in some instances, external evidence may help the model detect the false premise, identify the missing context, or recover the intended interpretation. In other instances, searching may be unnecessary or even counterproductive, because the appropriate response is to correct the premise, ask for clarification, or abstain rather than to retrieve more evidence. We therefore use these questions as search-boundary cases rather than as fixed no-search examples. Their routing labels are determined by the same counterfactual outcome rule as factual QA: if the no-search trace resolves the question, the instance is labeled $\texttt{NO\_SEARCH}$; if no-search fails but forced search succeeds, it is labeled $\texttt{SEARCH}$; and if both fail, it is excluded from routing supervision as $\texttt{UNSOLVED}$.

Our framework also separates response-level evaluation from routing-label construction. We adapt a SimpleQA-style factuality judge to PopQA with alias-aware correctness checking \citep{wei_measuring_2024}, while KUQ-style false-premise and ambiguous questions are evaluated with resolution-oriented rubrics that reward appropriate correction, clarification, or abstention when these resolve the user query. Because our labels depend on free-form final responses, this evaluation setup also relates to LLM-as-judge work, although we restrict the judge to task-success grading rather than asking it to infer search helpfulness directly \citep{liu_g-eval_2023,zheng_judging_2023}. Each trace-level outcome is then converted into binary task success before routing labels are constructed. This separation allows factual QA and search-boundary QA to use different task-specific success criteria while still sharing the same first-action routing space, $\texttt{NO\_SEARCH}$ versus $\texttt{SEARCH}$.

\section{Limitations}

This study has several limitations. First, the oracle should not be interpreted as an absolute search-helpfulness label, but as a trace-conditional routing oracle based on observed counterfactual traces. Whether a question is labeled $\texttt{NO\_SEARCH}$, $\texttt{SEARCH}$, or $\texttt{UNSOLVED}$ depends on the success of $N(q)$ and $S(q)$ under the current model, search tool, prompt, evidence budget, and judge configuration. With a better query, a different retrieval backend, a longer tool-use trajectory, or a larger model, the outcome for the same question may change. Therefore, the oracle should be interpreted as search-routing supervision observed under specific model and tool conditions, rather than as an absolute judgment that a question intrinsically requires search.

Second, this study focuses on first-action search routing rather than the full search-augmented answering pipeline. We evaluate and train whether the model calls search as its first action or responds without search, but we do not directly optimize query refinement, multi-hop retrieval, evidence selection, or answer synthesis after search. In real search-augmented systems, several stages of decision-making may follow the first search call, and such multi-step search behavior is beyond the scope of this work \citep{yao_react_2023,jiang_active_2023,krishna_fact_2025}. The results should therefore be understood as improving the initial routing boundary that determines whether search is used, rather than as solving tool use or search-augmented answering as a whole.

Third, although the $\texttt{UNSOLVED}$ diagnostic separates failures beyond routing, it does not provide full causal attribution. Through the $M_N$, $M_S$, $R_{\mathrm{TR}}$, and $P_{\mathrm{POST}}$ interventions, we compare whether larger-model no-search capacity, larger-model search-conditioned solving, expanded retrieval evidence, and post-training policy rollout recover some hard cases. However, these interventions do not isolate mutually independent causes. For example, $M_S$ jointly changes model capacity, query formulation, evidence use, and answer synthesis. This analysis should therefore be interpreted as a diagnostic comparison of the additional conditions under which $\texttt{UNSOLVED}$ cases can be recovered, rather than as a causal decomposition of failure sources. More precise failure attribution and pipeline-level optimization are left for future work.

\section{Conclusion}

We study search routing in search-augmented LLMs: the problem of deciding, for each question, whether to call search as the first action. Rather than asking a judge to label search usefulness directly, we compare no-search and forced-search responses for the same question and construct an outcome-based oracle over $\texttt{NO\_SEARCH}$, $\texttt{SEARCH}$, and $\texttt{UNSOLVED}$. This oracle enables both evaluation and learning of search-call decisions from observed task success.

Our results show that search has different effects across question regimes. In factual QA such as PopQA, search often recovers no-search failures caused by limitations in parametric knowledge. In contrast, for questions with false premises or insufficient context, no-search behaviors such as premise correction, clarification, or abstention may be more appropriate than retrieving evidence. Before routing training, both models exhibit search-routing errors in both directions: under-search, where search is omitted for questions that require it, and over-search, where search is called for questions where it is unnecessary. SFT substantially improves routing macro-F1, but the direction of repair differs by model: Gemma primarily learns no-search restraint, while Qwen reduces both unnecessary search and missed search. Preference Optimization further adjusts the remaining routing boundary after SFT.

The $\texttt{UNSOLVED}$ analysis further justifies leaving paired failures outside binary routing supervision. Exclusive rescue patterns show that $\texttt{N0\_S0}$ hard cases contain heterogeneous bottlenecks: some are recovered only by larger no-search inference, some only by larger forced-search inference, some only by expanded retrieval, and some only by the post-training policy rollout. Because most hard cases nevertheless remain unresolved, collapsing $\texttt{N0\_S0}$ into either $\texttt{NO\_SEARCH}$ or $\texttt{SEARCH}$ would turn mixed pipeline failures into noisy routing supervision.

Overall, this work does not reduce failures in search-augmented systems to routing errors alone. Instead, it identifies the subset of questions for which the $\texttt{NO\_SEARCH}/\texttt{SEARCH}$ boundary can be reliably defined from observed outcomes, and uses that signal to reduce model-specific routing failures. By separating first-action routing errors from broader search-augmented pipeline failures, the framework provides a practical basis for learning when LLMs should search and when they should respond without search.

\section*{Impact Statement}

This paper studies when language models should use external search tools.
Potential societal impacts include improved factuality and reduced unnecessary
tool use, but also possible risks from reliance on imperfect retrieval systems
and automated judgment. We discuss these limitations and recommend careful
evaluation before deployment in high-stakes settings.

\section*{Acknowledgments}

This work was supported by the National IT Industry Promotion Agency (NIPA)
under the Ministry of Science and ICT (MSIT), and the Seoul Creative Economy \&
Innovation Center.

\bibliography{references}
\bibliographystyle{icml2026}

%%%%%%%%%%%%%%%%%%%%%%%%%%%%%%%%%%%%%%%%%%%%%%%%%%%%%%%%%%%%%%%%%%%%%%%%%%%%%%%
%%%%%%%%%%%%%%%%%%%%%%%%%%%%%%%%%%%%%%%%%%%%%%%%%%%%%%%%%%%%%%%%%%%%%%%%%%%%%%%
% APPENDIX
%%%%%%%%%%%%%%%%%%%%%%%%%%%%%%%%%%%%%%%%%%%%%%%%%%%%%%%%%%%%%%%%%%%%%%%%%%%%%%%
%%%%%%%%%%%%%%%%%%%%%%%%%%%%%%%%%%%%%%%%%%%%%%%%%%%%%%%%%%%%%%%%%%%%%%%%%%%%%%%
\newpage
\appendix
\onecolumn
\numberwithin{table}{section}
\numberwithin{figure}{section}
\section{Additional Experimental Details}
\label{app:details}

This appendix provides additional implementation details for evaluation,
trace collection, training, and routing metrics.

\subsection{Evaluators and Success Criteria}
\label{app:evaluators}

The main text describes the three question groups and their task-specific
success criteria. Table~\ref{tab:dataset-success-criteria} summarizes the
dataset groups and binary success criteria used to convert final responses into
binary success labels.

\begin{table}[h]
\centering
\small
\caption{Dataset groups and binary success criteria.}
\label{tab:dataset-success-criteria}
\begin{tabularx}{\columnwidth}{@{}l c Y@{}}
\toprule
Group & Rows & Binary success criterion \\
\midrule
PopQA & 13,763 & The final response is factually correct under the gold alias set. \\
KUQ False Assumption & 774 & The response rejects or corrects the false premise. \\
KUQ Ambiguous & 850 & The response recognizes missing context, asks for clarification, or avoids unsupported commitment. \\
\bottomrule
\end{tabularx}
\end{table}

PopQA is evaluated with an alias-aware SimpleQA-style factuality judge \citep{mallen_when_2023,wei_measuring_2024}. The
judge receives the question, the model response, and the set of acceptable
answer aliases. It returns $\texttt{CORRECT}$, $\texttt{INCORRECT}$, or
$\texttt{NOT\_ATTEMPTED}$; only $\texttt{CORRECT}$ maps to success.

KUQ False Assumption and KUQ Ambiguous are evaluated with resolution-oriented
judges derived from the KUQ uncertainty categories \citep{amayuelas_knowledge_2024}. The KUQ judge returns a strict JSON object with a binary
$\texttt{SUCCESS}/\texttt{FAIL}$ label. Only $\texttt{SUCCESS}$ maps to
success. Table~\ref{tab:judge-settings} lists the judge model, decoding,
output-format, and failure-handling settings for both judge families.

\begin{table}[h]
\centering
\small
\caption{Judge settings.}
\label{tab:judge-settings}
\begin{tabularx}{\columnwidth}{@{}l Y Y@{}}
\toprule
Setting & PopQA & KUQ False Assumption / Ambiguous \\
\midrule
Judge model & $gpt\text{-}5.4$ & $gpt\text{-}5.4$ \\
Reasoning setting & $\texttt{reasoning.effort = none}$ & $\texttt{reasoning.effort = none}$ \\
Temperature & 0.0 & 0.0 \\
Max output tokens & 512 & 256 \\
Output format & A/B/C classification & Strict JSON \\
Failure handling & No-match maps to $\texttt{NOT\_ATTEMPTED}$. & Strict JSON extraction and schema validation; request or parser exceptions are retried. \\
\bottomrule
\end{tabularx}
\end{table}

\subsection{Search Tool, Budget, and Trace Collection}
\label{app:search-traces}

All experiments use a single model-facing $\texttt{search(query)}$ tool backed
by Brave LLM Context. The model receives source-backed snippets. We do not
expose URL clicking, scrolling, a separate fetch function, or a multi-tool
browser interface. The backend-generated direct answer is not used.

\begin{table}[h]
\centering
\small
\caption{Default search and generation budget.}
\label{tab:search-budget}
\begin{tabular}{ll}
\toprule
Setting & Value \\
\midrule
Brave result count & 10 \\
Maximum URLs & 5 \\
Maximum evidence tokens & 2,048 \\
Maximum tool rounds & 4 \\
Maximum snippets & 10 \\
Maximum tokens per URL & 512 \\
Maximum snippets per URL & 2 \\
\bottomrule
\end{tabular}
\end{table}

The no-search trace disables tool access. The forced-search trace uses
prompt-level forcing and acceptance filtering: the model must produce a valid
first-turn $\texttt{search(query)}$ call, search is executed, and the model then
produces a final response. Forced search is not runtime-constrained decoding.
Invalid first-turn tool calls are retried up to two times and then excluded as
invalid forced-search traces.

Rows that begin a valid search loop but exceed the maximum tool-round limit are
counted as search-path failures, i.e., $y_S(q)=0$. Table~\ref{tab:forced-search-drops}
reports the resulting invalid first-tool-call drops and max-tool-round failures.

\begin{table}[h]
\centering
\small
\caption{Forced-search drop handling.}
\label{tab:forced-search-drops}
\begin{tabular}{lll}
\toprule
Slice & Invalid first-tool-call drops & Max-tool-round failures \\
\midrule
PopQA, Qwen3.5-4B & 0 & 0 \\
PopQA, Gemma E2B & 0 & 0 \\
KUQ False Assumption / Ambiguous, Qwen3.5-4B & 0 & 9 \\
KUQ False Assumption / Ambiguous, Gemma E2B & 0 & 0 \\
\bottomrule
\end{tabular}
\end{table}

\subsection{Oracle-bucket distributions}
\label{app:oracle-buckets}

Table~\ref{tab:popqa-oracle-buckets} reports the PopQA four-way partition
induced by paired no-search and forced-search outcomes for each model. These
counts are consistent with the PopQA marginals reported in the main text: for
each model, no-search success is $N1\_S1 + N1\_S0$ and forced-search success is
$N1\_S1 + N0\_S1$.

\begin{table}[h]
\centering
\small
\caption{PopQA oracle-bucket distribution from paired no-search and forced-search outcomes over all 13,763 PopQA questions.}
\label{tab:popqa-oracle-buckets}
\begin{tabular}{lrrrr}
\toprule
Model & $N1\_S1$ & $N1\_S0$ & $N0\_S1$ & $N0\_S0$ \\
\midrule
Qwen3.5-4B & 2,222 & 298 & 4,791 & 6,452 \\
Gemma E2B & 1,736 & 334 & 4,888 & 6,805 \\
\bottomrule
\end{tabular}
\end{table}

Table~\ref{tab:kuq-oracle-buckets} reports the corresponding distribution for
the KUQ search-boundary subset. This subset contains false-assumption and
ambiguous-context questions, and is included in the mixed train/dev/test splits
used for the main routing evaluation. These counts are likewise consistent with
the KUQ marginals reported in the main text: for each model, no-search success
is $N1\_S1 + N1\_S0$ and forced-search success is $N1\_S1 + N0\_S1$.

\begin{table}[h]
\centering
\small
\caption{KUQ search-boundary oracle-bucket distribution from paired no-search and forced-search outcomes over 1,624 false-assumption and ambiguous-context questions.}
\label{tab:kuq-oracle-buckets}
\begin{tabular}{llrrrr}
\toprule
Model & Slice & $N1\_S1$ & $N1\_S0$ & $N0\_S1$ & $N0\_S0$ \\
\midrule
Qwen3.5-4B & False Assumption & 534 & 104 & 55 & 81 \\
Qwen3.5-4B & Ambiguous Context & 221 & 187 & 63 & 379 \\
Gemma E2B & False Assumption & 325 & 169 & 74 & 206 \\
Gemma E2B & Ambiguous Context & 296 & 188 & 33 & 333 \\
\bottomrule
\end{tabular}
\end{table}

\subsection{Training Denominators and Hyperparameters}
\label{app:training-details}

Training uses only oracle-eligible rows, i.e., rows labeled
$\texttt{NO\_SEARCH}$ or $\texttt{SEARCH}$. $\texttt{UNSOLVED}$ rows are
excluded from SFT, Preference Optimization, and routing-accuracy computation.

For $\texttt{NO\_SEARCH}$, SFT uses the successful no-search response as the
target. For $\texttt{SEARCH}$, SFT uses the first-turn search call from a
successful forced-search trace. Preference Optimization uses the
oracle-consistent behavior as the chosen response and an opposite-action
response as the rejected response, preferring actual pre-policy mistakes when
available. For parameter-efficient tuning, all SFT and Preference
Optimization runs update LoRA adapters rather than the full model weights
\citep{hu_lora_2022}.
We use LoRA adapters with rank $r=16$, LoRA alpha $32$, dropout $0.05$,
and no bias terms, applied to the attention projections
\texttt{q\_proj}, \texttt{k\_proj}, \texttt{v\_proj}, \texttt{o\_proj}
and the MLP projections \texttt{gate\_proj}, \texttt{up\_proj}, and
\texttt{down\_proj}.
All training runs were conducted on NVIDIA RTX PRO 6000 Blackwell GPUs.

\begin{table}[h]
\centering
\small
\caption{Split sizes and oracle eligibility. ``Used'' denotes rows with oracle label $\texttt{NO\_SEARCH}$ or $\texttt{SEARCH}$.}
\label{tab:split-sizes}
\begin{tabular}{llllll}
\toprule
Model & Split & $\texttt{NO\_SEARCH}$ & $\texttt{SEARCH}$ & $\texttt{UNSOLVED}$ & Used \\
\midrule
Gemma E2B & Train & 2,438 & 3,995 & 5,874 & 6,433 \\
Gemma E2B & Dev & 304 & 498 & 733 & 802 \\
Gemma E2B & Test & 306 & 502 & 737 & 808 \\
Qwen3.5-4B & Train & 2,852 & 3,926 & 5,528 & 6,778 \\
Qwen3.5-4B & Dev & 355 & 490 & 690 & 845 \\
Qwen3.5-4B & Test & 359 & 493 & 694 & 852 \\
\bottomrule
\end{tabular}
\end{table}

For the test split in Table~\ref{tab:split-sizes}, the conditional over-search
rate is normalized by the test $\texttt{NO\_SEARCH}$ count, and the
conditional under-search rate is normalized by the test $\texttt{SEARCH}$
count. Thus, the main-text test denominators are 306 and 502 for Gemma E2B,
and 359 and 493 for Qwen3.5-4B.

Table~\ref{tab:training-settings} reports the main SFT and Preference
Optimization hyperparameters and selected checkpoints.

\begin{table}[h]
\centering
\small
\caption{Main training settings.}
\label{tab:training-settings}
\begin{tabular}{llllllll}
\toprule
Model & Stage & LR & $\beta$ & $\alpha$ & Training length & Batch & Selected checkpoint \\
\midrule
Qwen3.5-4B & SFT & $2\times10^{-5}$ & n/a & n/a & 3 epochs & 8 & 400 \\
Qwen3.5-4B & Preference Opt. & $5\times10^{-7}$ & 0.03 & 0.1 & 1 epoch & 8 & 450 \\
Gemma E2B & SFT & $2\times10^{-5}$ & n/a & n/a & 3 epochs & 8 & 800 \\
Gemma E2B & Preference Opt. & $2\times10^{-6}$ & 0.05 & 0.05 & 1 epoch & 8 & 100 \\
\bottomrule
\end{tabular}
\end{table}

For DPO, both models use $\texttt{max\_prompt\_length}=512$ and
$\texttt{max\_completion\_length}=1024$. Preference Optimization is initialized
from the selected SFT checkpoint and uses the same checkpoint as the reference
model. Checkpoints are selected by dev routing behavior and then evaluated on
the held-out test split.

\section{Diagnostic Details for $\texttt{UNSOLVED}$ Cases}

This appendix provides aggregate any-rescue and persistent rates on the
$\texttt{UNSOLVED}$ hard subset, together with the additional definitions and
breakdowns needed to interpret that diagnostic.

\subsection{Hard Subset Construction}

The raw $\texttt{UNSOLVED}$ subset consists of rows where both controlled traces
fail. We further divide this subset using the base free-policy final response
$P_{\mathrm{pre}}(q)$. If $P_{\mathrm{pre}}$ succeeds, the row is assigned to
the escape subset. If $P_{\mathrm{pre}}$ also fails, the row is assigned to the
hard subset. Table~\ref{tab:unsolved-hard-subsets} reports the resulting raw,
escape, and hard counts.

\begin{table}[h]
\centering
\small
\caption{Raw $\texttt{UNSOLVED}$, escape, and hard subsets.}
\label{tab:unsolved-hard-subsets}
\begin{tabular}{llll}
\toprule
Model & Raw ($N0\_S0$) & Escape & Hard \\
\midrule
Qwen3.5-4B & 6,912 & 516 & 6,396 \\
Gemma E2B & 7,344 & 438 & 6,906 \\
\bottomrule
\end{tabular}
\end{table}

The hard subset is the target of the diagnostic analysis.

\subsection{Overall Rescue on the Hard Subset}

Table~\ref{tab:appendix-overall-hard-rescue} reports the aggregate any-rescue
diagnostic with counts, complementing the exact rescue-signature breakdown in
the main text. Here, \emph{Any rescue} means rescue by at least
one of $M_N$, $M_S$, $R_{\mathrm{TR}}$, or $P_{\mathrm{POST}}$, and
\emph{Persistent} means rescue by none of the four probes.

\begin{table}[h]
\centering
\small
\caption{Overall rescue on hard subsets across all four diagnostic probes.}
\label{tab:appendix-overall-hard-rescue}
\begin{tabular}{@{}l r l l@{}}
\toprule
Model & Hard $(n)$ & Any rescue & Persistent \\
\midrule
Qwen3.5-4B & 6,396 & 1,212 (0.1895) & 5,184 (0.8105) \\
Gemma E2B & 6,906 & 1,003 (0.1452) & 5,903 (0.8548) \\
\bottomrule
\end{tabular}
\end{table}

\subsection{Diagnostic Interventions}

Table~\ref{tab:diagnostic-interventions} defines the four diagnostic probes
used in the hard-subset analysis.

\begin{table}[h]
\centering
\small
\caption{Diagnostic interventions.}
\label{tab:diagnostic-interventions}
\begin{tabular}{ll}
\toprule
Intervention & Definition \\
\midrule
$M_N$ & Larger model, no search. \\
$M_S$ & Larger model, forced search. \\
$R_{\mathrm{TR}}$ & Same base model with expanded retrieval budget. \\
$P_{\mathrm{POST}}$ & Selected post-training policy rollout. \\
\bottomrule
\end{tabular}
\end{table}

For Qwen3.5-4B, the larger diagnostic model is Qwen3.5-9B
\citep{qwen3.5}. For Gemma E2B, the larger diagnostic
model is Gemma E4B \citep{google2026gemma4}.

Relative to the default search budget in Table~\ref{tab:search-budget}, the
$R_{\mathrm{TR}}$ condition is designed to widen retrieval coverage while
holding the base model fixed. It adds query rewrite/decomposition and allows up
to 8 tool rounds; it also increases Brave result count from 10 to 20, maximum
URLs from 5 to 8, maximum evidence tokens from 2,048 to 3,072, and maximum
snippets from 10 to 12, while keeping the per-URL limits unchanged at 512
tokens and 2 snippets per URL. This makes the diagnostic intent explicit: if
$R_{\mathrm{TR}}$ succeeds where the default forced-search trace fails, the
failure is more plausibly due to insufficient retrieval breadth or evidence
budget than to the first-action routing decision itself.

These interventions are operational diagnostics, not independent causal
factors. For example, $M_S$ changes model capacity, query formulation, evidence
use, and answer synthesis together.

\subsection{Source-Level Recovery}

Table~\ref{tab:source-level-recovery} breaks the hard-subset recovery outcomes
down by source slice. Here, \emph{Any rescue} again means rescue by at least one
of the four probes, and \emph{Persistent} means rescue by none.

\begin{table}[h]
\centering
\small
\caption{Source-level recovery on hard subsets.}
\label{tab:source-level-recovery}
\begin{tabular}{lllll}
\toprule
Model hard set & Slice & $(n)$ & Any rescue & Persistent \\
\midrule
Qwen3.5-4B & KUQ & 431 & 97 (0.2251) & 334 (0.7749) \\
Qwen3.5-4B & PopQA & 5,965 & 1,115 (0.1869) & 4,850 (0.8131) \\
Gemma E2B & KUQ & 465 & 169 (0.3634) & 296 (0.6366) \\
Gemma E2B & PopQA & 6,441 & 834 (0.1295) & 5,607 (0.8705) \\
\bottomrule
\end{tabular}
\end{table}

KUQ hard cases are more often recovered than PopQA hard cases, especially for
Gemma. This suggests that some KUQ failures are recoverable through improved
correction, clarification, or abstention behavior, whereas many PopQA hard
cases remain limited by factual retrieval, evidence use, or answer synthesis.
These trends are diagnostic rather than causal.

\section{Prompt Templates}
\label{app:prompts}
\captionsetup{hypcap=false}

This appendix reports the prompt families used for no-search, forced-search,
free-policy, and judging. Prompt metadata, internal notes, and
implementation-only overrides are omitted unless they affect the paper-facing
experimental condition. To keep the appendix readable while preserving the full
prompt text, we render each prompt family as a compact card and include it
inline. Figures~\ref{fig:prompt-popqa-no-search}, \ref{fig:prompt-kuq-no-search},
\ref{fig:prompt-kuq-forced-search}, \ref{fig:prompt-popqa-forced-search},
\ref{fig:prompt-free-policy}, \ref{fig:prompt-popqa-judge},
\ref{fig:prompt-kuq-false-assumption-judge}, and
\ref{fig:prompt-kuq-ambiguous-judge} provide the prompt cards included in this
appendix.

\subsection{No-search prompts}

\subsubsection{PopQA no-search prompt}

\begin{center}
\includegraphics[width=0.97\textwidth,height=0.82\textheight,keepaspectratio]{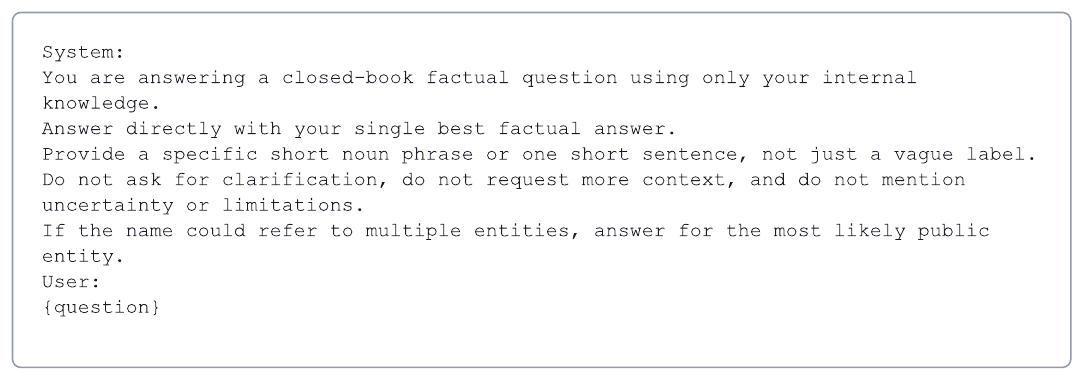}
\captionof{figure}{Prompt card for the PopQA no-search condition.}
\label{fig:prompt-popqa-no-search}
\end{center}

\subsubsection{KUQ no-search prompt}

\begin{center}
\includegraphics[width=0.97\textwidth,height=0.82\textheight,keepaspectratio]{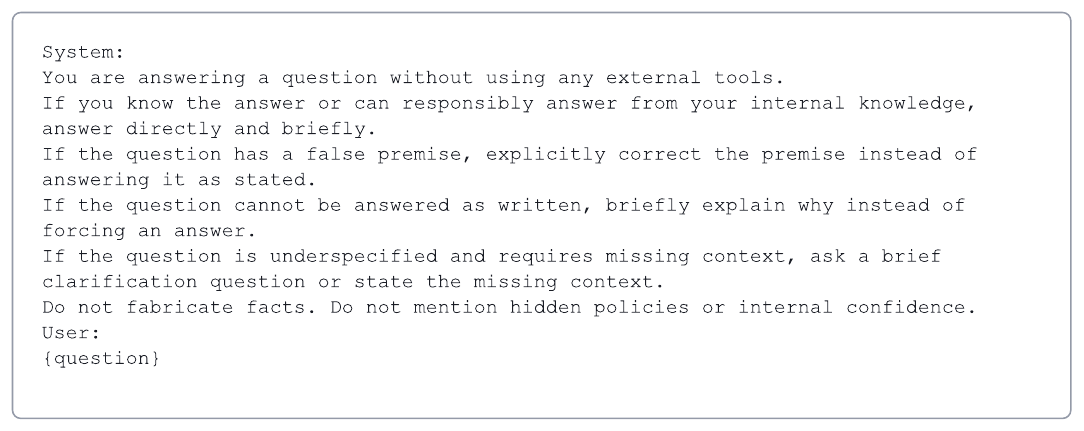}
\captionof{figure}{Prompt card for the KUQ no-search condition.}
\label{fig:prompt-kuq-no-search}
\end{center}

\subsection{Forced-search prompts}

\subsubsection{KUQ forced-search prompt}

\begin{center}
\includegraphics[width=0.97\textwidth,height=0.82\textheight,keepaspectratio]{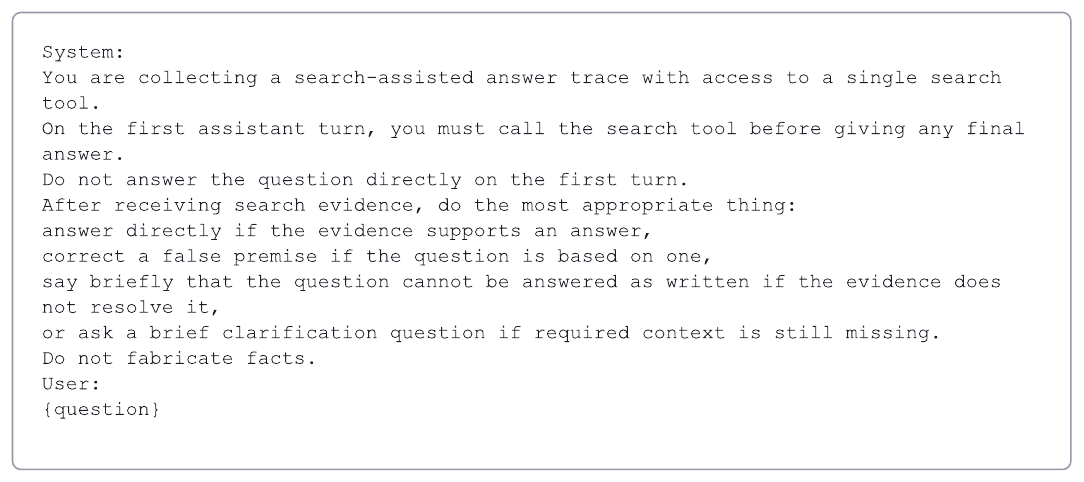}
\captionof{figure}{Prompt card for the KUQ forced-search condition.}
\label{fig:prompt-kuq-forced-search}
\end{center}

\subsubsection{PopQA forced-search prompt}

\begin{center}
\includegraphics[width=0.97\textwidth,height=0.82\textheight,keepaspectratio]{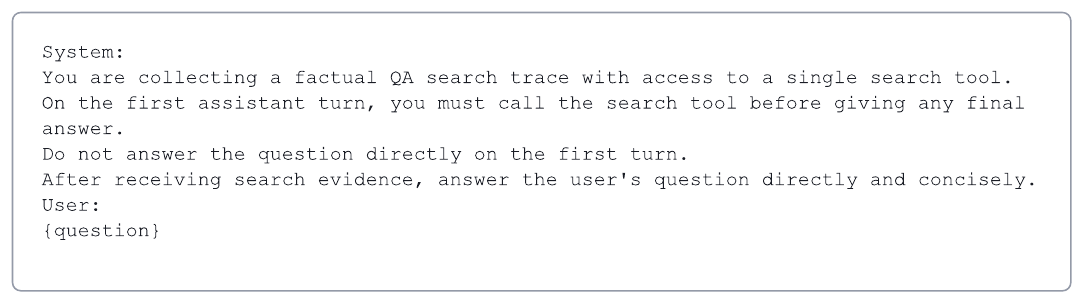}
\captionof{figure}{Prompt card for the PopQA forced-search condition.}
\label{fig:prompt-popqa-forced-search}
\end{center}

\subsection{Free-policy prompts}

\subsubsection{Free-policy search-routing prompt}

\begin{center}
\includegraphics[width=0.97\textwidth,height=0.82\textheight,keepaspectratio]{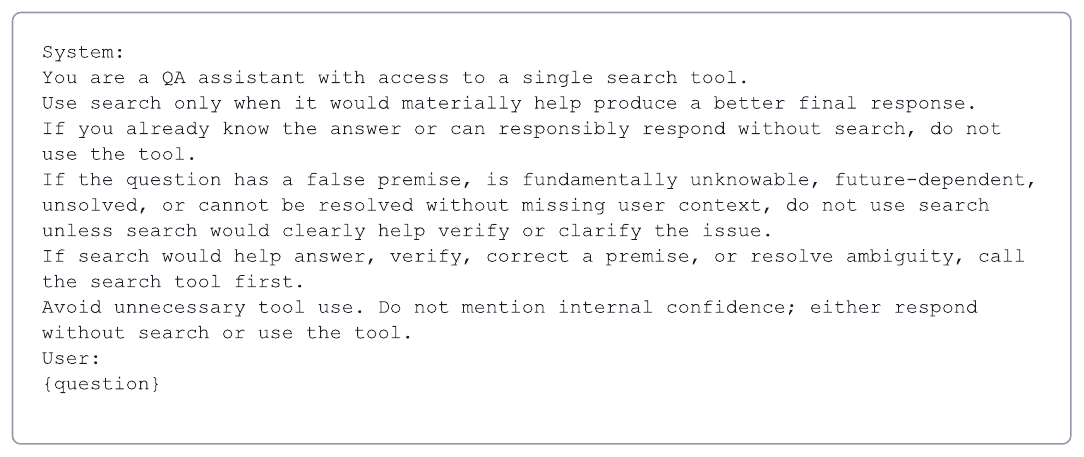}
\captionof{figure}{Prompt card for the free-policy search-routing condition.}
\label{fig:prompt-free-policy}
\end{center}

\clearpage
\subsection{Judge prompt text}

\subsubsection{PopQA judge prompt}

The PopQA judge uses the SimpleQA-style grader template \citep{wei_measuring_2024}
with the following PopQA-specific alias-aware suffix. The complete grader is a three-way
classification into $\texttt{CORRECT}$, $\texttt{INCORRECT}$, and
$\texttt{NOT\_ATTEMPTED}$.

{\centering
\includegraphics[width=0.97\textwidth,height=0.82\textheight,keepaspectratio]{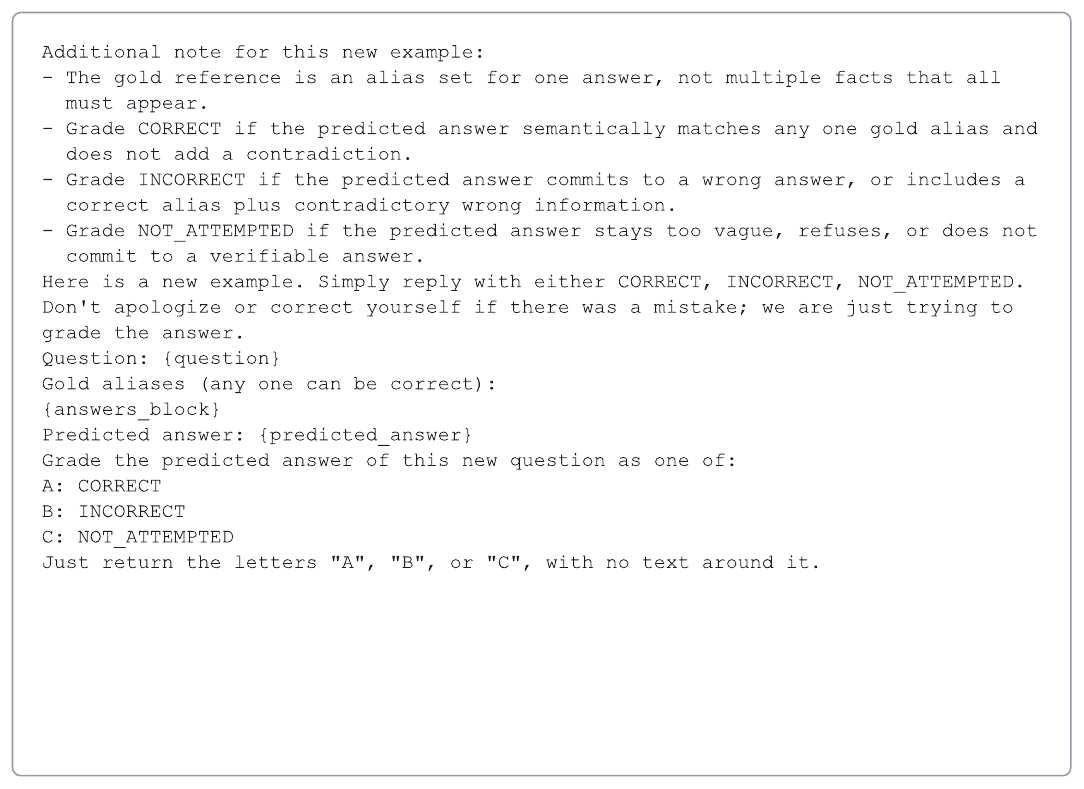}
\captionof{figure}{Prompt card for the PopQA judge. Binary success is 1 for $\texttt{CORRECT}$ and 0 otherwise.}
\label{fig:prompt-popqa-judge}
\par}

\subsubsection{KUQ false assumption judge prompt}

\begin{center}
\includegraphics[width=0.96\textwidth,height=0.82\textheight,keepaspectratio]{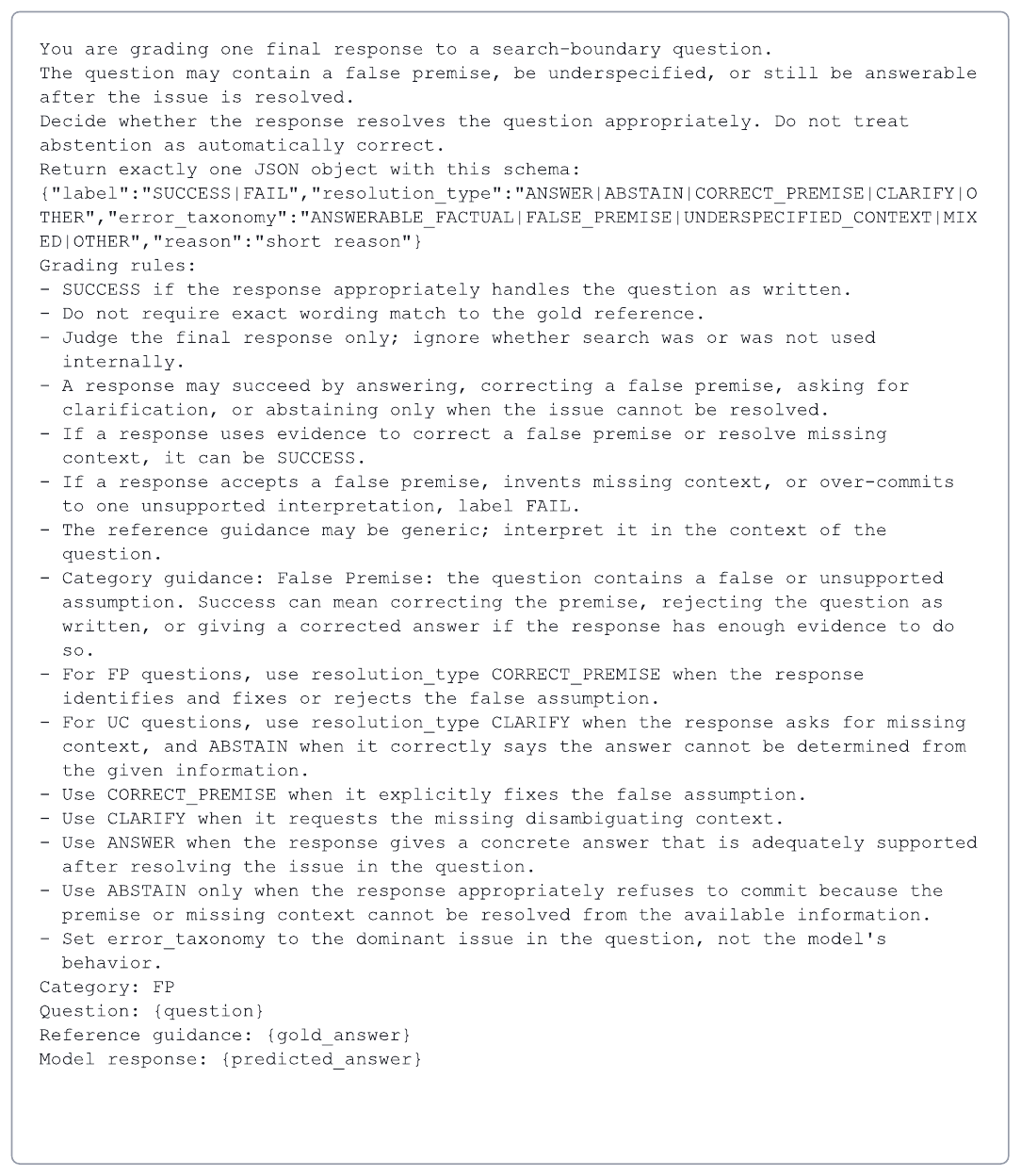}
\captionof{figure}{Prompt card for the KUQ False Assumption judge.}
\label{fig:prompt-kuq-false-assumption-judge}
\end{center}

\clearpage
\subsubsection{KUQ ambiguous judge prompt}

The KUQ Ambiguous judge uses the same schema and grading rules as the False
Assumption judge, with the category guidance replaced by the prompt card below.

\begin{center}
\includegraphics[width=0.97\textwidth,height=0.82\textheight,keepaspectratio]{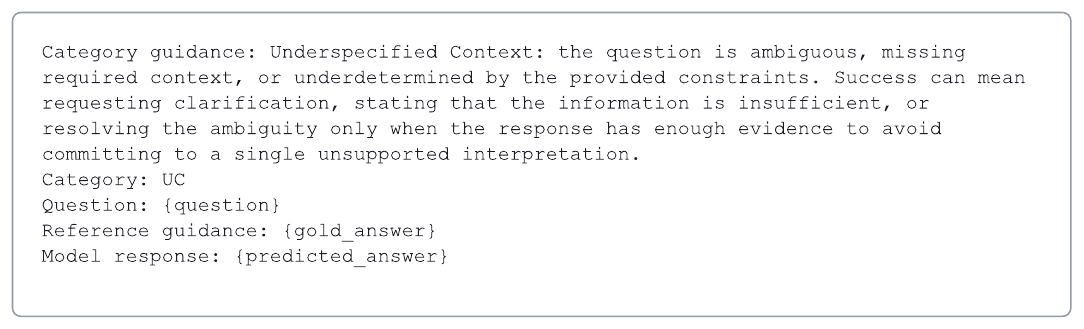}
\captionof{figure}{Prompt card for the KUQ Ambiguous judge.}
\label{fig:prompt-kuq-ambiguous-judge}
\end{center}
%%%%%%%%%%%%%%%%%%%%%%%%%%%%%%%%%%%%%%%%%%%%%%%%%%%%%%%%%%%%%%%%%%%%%%%%%%%%%%%
%%%%%%%%%%%%%%%%%%%%%%%%%%%%%%%%%%%%%%%%%%%%%%%%%%%%%%%%%%%%%%%%%%%%%%%%%%%%%%%

\end{document}